\documentclass[runningheads]{llncs}
\usepackage{graphicx}
\usepackage{booktabs}
\usepackage{multirow}
\begin{document}
\title{Using ChatGPT for Entity Matching}
%
%
\author{Ralph Peeters\orcidID{0000-0003-3174-2616} \and
Christian Bizer\orcidID{0000-0003-2367-0237}}
\authorrunning{Peeters and Bizer}
%
\institute{Data and Web Science Group, University of Mannheim, Germany
\email{\{ralph.peeters,christian.bizer\}@uni-mannheim.de}}
\maketitle              
\begin{abstract}
Entity Matching is the task of deciding if two entity descriptions refer to the same real-world entity. State-of-the-art entity matching methods often rely on fine-tuning Transformer models such as BERT or RoBERTa. Two major drawbacks of using these models for entity matching are that (i) the models require significant amounts of fine-tuning data for reaching a good performance and (ii) the fine-tuned models are not robust concerning out-of-distribution entities. In this paper, we investigate using ChatGPT for entity matching as a more robust, training data-efficient alternative to traditional Transformer models.  We perform experiments along three dimensions: (i) general prompt design, (ii) in-context learning, and (iii) provision of higher-level matching knowledge. We show that ChatGPT is competitive with a fine-tuned RoBERTa model, reaching a zero-shot performance of 82.35\% F1 on a challenging matching task on which RoBERTa requires 2000 training examples for reaching a similar performance. Adding in-context demonstrations to the prompts further improves the F1 by up to 7.85\% when using similarity-based example selection. Always using the same set of 10 handpicked demonstrations leads to an improvement of 4.92\%  over the zero-shot performance. Finally, we show that ChatGPT can also be guided by adding higher-level matching knowledge in the form of rules to the prompts. Providing matching rules leads to similar performance gains as providing in-context demonstrations.

\end{abstract}

\section{Introduction}

Entity matching is the task of discovering entity descriptions in different data sources that refer to the same real-world entity~\cite{christophides_end--end_2020}. While
early matching systems relied on manually defined matching rules,
supervised machine learning methods have become the foundation
of most entity matching systems~\cite{christophides_end--end_2020} since the 2000s. This trend
was reinforced by the success of neural networks~\cite{BarlaugNeural2021} and
today most state-of-the art matching systems rely on pre-trained language models (PLMs), such as BERT or RoBERTa~\cite{liDeepEntityMatching2020,peetersSupervisedContrastiveLearning2022a,peeters2023wdc}.

The downsides of using PLMs for entity matching are that (i) PLMs need a lot of task-specific training examples for fine-tuning and (ii) they are not very robust concerning unseen entities that were not part of the training data~\cite{akbarian2022probing,peeters2023wdc}.

Large autoregressive language models (LLMs)~\cite{zhao2023survey} such as GPT, ChatGPT, PaLM, or BLOOM have the potential to address both of these shortcomings. Due to being pre-trained on huge amounts of text as well as due to emergent effects resulting from the model size~\cite{wei2022emergent}, LLMs often have a better zero-shot performance compared to PLMs such as BERT and are also more robust concerning unseen examples~\cite{brown2020language}. 
Initial research on exploring the potential of LLMs for data wrangling tasks was conducted by Narayan et al.~\cite{foundationalWrangleVLDB2022} using the GPT-3 LLM.
This paper builds on the results of Narayan et al. and extends them with the following contributions:

\begin{enumerate}
    \item We are the first to systematically evaluate the performance of ChatGPT (gpt3.5-turbo-0301) on the task of entity matching, while Narayan et al. applied the earlier GPT-3 model (text-davinci-002).
    \item We systematically compare various prompt design options for entity matching while Narayan et al. tested only two designs. 
    \item We extend the results of Narayan et al. on in-context learning for entity matching  by introducing a similarity-based method for selecting demonstrations from a pool of training examples. 
    Furthermore, we analyze the impact of in-context learning on the costs (usage-fees) charged for running entity matching prompts against the OpenAI API.  
    \item We are the first to experiment with the addition of higher-level matching knowledge to prompts as an alternative to in-context demonstrations. We show that guiding the model by stating higher-level matching rules can lead to the same positive effect as providing in-context examples.
\end{enumerate}


\section{Experimental Setup}
\label{Sec_Setup}

Narayan et al.~\cite{foundationalWrangleVLDB2022} have measured the performance of GPT-3 using a range of well-known entity matching benchmark datasets~\cite{primpeliProfilingEntityMatching2020}. All of these datasets are available on the Web for quite some time and are widely discussed in various papers and on various webpages. Thus, it is very likely that the training data of GPT-3 and ChatGPT contains information about these benchmarks which could give the language models an advantage. In order to eliminate this potential of leaking information about the test sets, we use the WDC Products~\cite{peeters2023wdc} benchmark which has been published in December 2022 and is therefore newer than the training data of the tested models. 

In order to understand the potential of LLMs for challenging entity matching use cases, we use a difficult variant of the WDC Products benchmark for the experiments which contains 80\% corner-cases (hard positives and hard negatives). 
The types of products that are contained in our evaluation dataset range from computers and electronics over bike parts to general tools and thus examplify different product categories. 
The products are described by the attributes \textit{brand}, \textit{title}, \textit{description} and \textit{price}. 
In order to keep the costs of running benchmark experiments against the OpenAI API in an acceptable range, we down-sample the WDC Products benchmark to 50 products and retain the high ratio of corner-cases by using the original benchmark creation code. Table \ref{tab:datasets} shows statistics about the original and down-sampled versions of the benchmark. 

\begin{table}[]
\centering
\caption{Statistics of the WDC Products benchmark datasets.}
\label{tab:datasets}
\resizebox{\textwidth}{!}{%
\begin{tabular}{@{}lcccc@{}}
\toprule
Dataset Type          & Purpose                      & \# Pairs & \# Pos & \# Neg \\ \midrule
Original Validation  & RoBERTa baseline                            & 4,500    & 500    & 4,000  \\
Original Training   & RoBERTa baseline                           & 19,835   & 8,741  & 11,364 \\ \midrule
Sampled Validation   & Evaluation of prompts & 433      & 50     & 383    \\
Sampled Training   & In-context sample selection  & 2025     & 898    & 1,127  \\ 
\bottomrule
\end{tabular}%
}
\end{table}

\textbf{API Calls and Costs:} We use the down-sampled validation set to report the impact of the various prompt design decisions and the down-sampled training set as a source of in-context demonstrations for the corresponding experiments. Thus, one evaluation run results in 433 API calls to the OpenAI API. For all experiments we use the ChatGPT version \textit{gpt3.5-turbo-0301} and set the temperature parameter to 0 to make experiments reproducible as stated in the OpenAI guidelines. We further track the cost associated with each pass of the validation set by using the Tiktokenizer\footnote{https://github.com/dqbd/tiktokenizer} python package to calculate the cost associated with each prompt and corresponding ChatGPT answer.

\begin{figure}
\includegraphics[width=\textwidth]{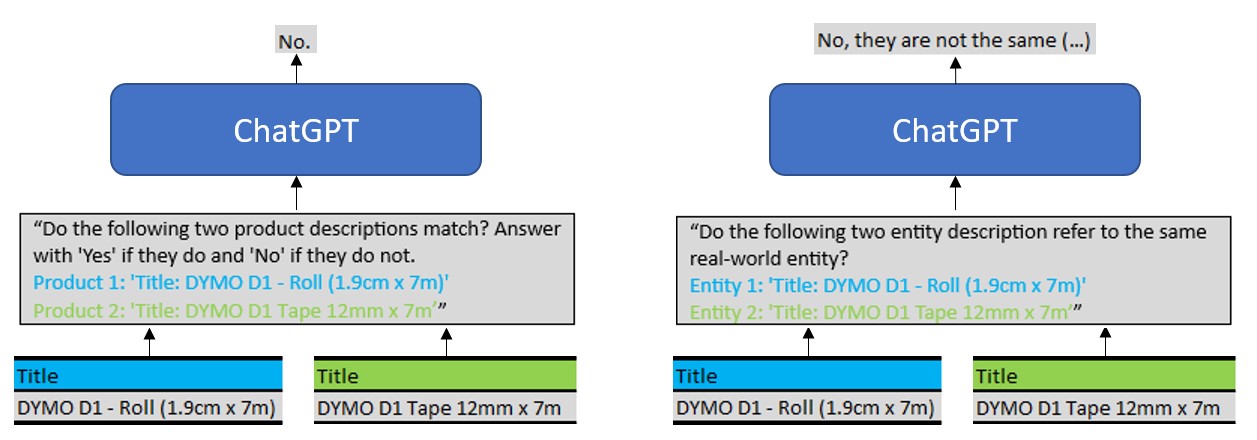}
\caption{Examples of prompt designs and product offer serializations.} \label{fig:general-prompts}
\end{figure}

\textbf{Serialization:} For the serialization of product offers into prompts, we follow related work~\cite{foundationalWrangleVLDB2022} and serialize each offer as a string with pre-pended attribute names. Figure \ref{fig:general-prompts} shows examples of this serialization practice for a pair of product offers and the attribute \textit{title}. 

\textbf{Evaluation:} The responses gathered from the model are natural language text. In order to decide if a response refers to a positive matching decision regarding a pair of product offers, we apply simple pre-processing to the answer and subsequently parse for the word \textit{yes}. In any other case we assume the model decides on not matching. This rather simple approach turns out to be surprisingly effective as the high recall values in Table \ref{tab:general-prompt-design} and manual inspection of the answers suggest. This approach has also been used by Narayan et al.~\cite{foundationalWrangleVLDB2022}. 

\textbf{Replicability:} All data and code used in this paper are available at the project github\footnote{https://github.com/wbsg-uni-mannheim/MatchGPT} meaning that all experiments can be replicated. In addition, we contributed the down-sampled datasets and three selected prompts to the OpenAI evals\footnote{https://github.com/openai/evals/blob/main/evals/registry/evals/product-matching.yaml} library.

\section{General Prompt Design}
\label{Sec_PromptDesign}

Designing the prompt input to large language models to convey the task description, input, as well as additional information is one of the main challenges for achieving good results~\cite{promptingSurvey2023}. Careful prompt design is important as it can have a large impact on the overall task performance~\cite{foundationalWrangleVLDB2022,zhao2021calibrate}. In this Section, we experiment with various prompt designs for ChatGPT in a zero-shot setting: We describe the task to the model and ask for a matching decision for each of the examples in our validation set. The prompt designs that we use can be categorized as follows and are illustrated by the example prompts in Figure \ref{fig:general-prompts}:

\begin{itemize}
    \item \textbf{General}: These prompts describe the task as the matching of entity descriptions to real-world entities. The product offers are referred to as \textit{entities}. An example of a \textit{general} prompt is the right prompt in Figure \ref{fig:general-prompts}.
    \item \textbf{Domain}: The domain-specific prompts describe the task as matching of product descriptions and refers to the examples as \textit{product offers}. An example of this type of prompt is the left prompt in Figure \ref{fig:general-prompts}.
    \item \textbf{Complex}: Prompts in this category use more complex language, specifically they use the formulations "refer to the same real-world product" or "refer to the same real-world entity". An example is the right prompt in Figure \ref{fig:general-prompts}.
    \item \textbf{Simple}: This type of prompt uses less complex language and replaces the formulations from \textit{Complex} with a simple "match". An example is the left prompt in Figure \ref{fig:general-prompts}.
    \item \textbf{Free}: This category reflects prompts that do not restrict the models answers in any way. An example is the right prompt in Figure \ref{fig:general-prompts}.
    \item \textbf{Forced}: In contrast to \textit{Free}, these kinds of prompts explicitly tell the model to answer the stated question with "Yes" and "No". An example is the left prompt in Figure \ref{fig:general-prompts}
    \item \textbf{Attributes}: We vary using the three attributes \textit{brand} (B), \textit{title} (T) and \textit{price} (P) in the combinations T, BT and BTP when serializing product offers into single strings.
\end{itemize}

\begin{table}[]
\centering
\caption{Results of the general prompt design experiment with associated cost.}
\label{tab:general-prompt-design}
\resizebox{0.8\columnwidth}{!}{%
\begin{tabular}{@{}lccccc@{}}
\toprule
Prompt             & P              & R               & F1             & \begin{tabular}[c]{@{}c@{}}$\Delta$ F1\end{tabular} & \begin{tabular}[c]{@{}c@{}}cost (¢)\\ per pair\end{tabular} \\ \midrule
general-complex-free-T    & 49.50          & \textbf{100.00} & 66.23          & -                                                  & 0.11                                                        \\
general-simple-free-T     & 70.00          & 98.00           & 81.67          & 15.44                                              & 0.10                                                        \\
general-complex-forced-T  & 63.29          & \textbf{100.00} & 77.52          & 11.29                                              & 0.14                                                        \\
general-simple-forced-T   & 75.38          & 98.00           & 85.22          & 18.99                                              & 0.13                                                        \\
general-simple-forced-BT  & 79.66          & 94.00           & \textbf{86.24} & 20.01                                              & 0.13                                                        \\
general-simple-forced-BTP & 71.43          & 70.00           & 70.70          & 4.47                                               & 0.13                                                        \\ \midrule
domain-complex-free-T     & 71.01          & 98.00           & 82.35          & 16.12                                              & 0.11                                                        \\
domain-simple-free-T      & 61.25          & 98.00           & 75.38          & 9.15                                               & 0.10                                                        \\
domain-complex-forced-T   & 71.01          & 98.00           & 82.35          & 16.12                                              & 0.14                                                        \\
domain-simple-forced-T    & 74.24          & 98.00           & 84.48          & 18.25                                              & 0.13                                                        \\
domain-simple-forced-BT   & 76.19          & 96.00           & 84.96          & 18.73                                              & 0.13                                                        \\
domain-simple-forced-BTP  & 54.54          & 84.00           & 66.14          & -0.09                                              & 0.13                                                        \\ \midrule
Narayan-complex-T    & 85.42          & 82.00           & 83.67          & 17.44                                              & 0.10                                                        \\
Narayan-simple-T     & \textbf{92.86} & 78.00           & 84.78          & 18.55                                              & 0.10                                                        \\ \bottomrule
\end{tabular}%
}
\end{table}

Table \ref{tab:general-prompt-design} shows the results of the experiments with general prompt designs and associated average cost for querying a single pair with each design. The recall values for all prompts with ChatGPT are equal to or above 98\% which suggests that, in combination with the lower precision values, the model is inclined to overestimating matching pairs in these cases. Interestingly, the prompt design of Narayan et al.~\cite{foundationalWrangleVLDB2022} that we also evaluate using ChatGPT, conversely results in a more balanced precision and recall, the latter being significantly lower than the ones we observe for our prompts. The main difference between Narayan et al.'s and our prompts is that they provide the examples to be matched before the task description while we do it the other way around. Comparing the F1 values of our general and domain-specific prompts with ChatGPT, three patterns emerge: (i) Formulating the prompt with domain-specific wording leads to generally more stable results, (ii) Using simpler language works better than more complex wording in all but one case and (iii) forcing the model to answer with a short "Yes" or "No" leads to a significant increase in every scenario. While the addition of brand information increases F1 by up to 1\% F1 percentage point, adding the price as well leads to a significant decrease in performance likely due to the format and currency of the prices not being normalized in the dataset.


\begin{table}[]
\centering
\caption{Results of the baseline experiments.}
\label{tab:baselines}
\resizebox{\columnwidth}{!}{%
\begin{tabular}{@{}llccccc@{}}
\toprule
Model                               & Configuration                                                                             & P     & R      & F1    & \begin{tabular}[c]{@{}c@{}}$\Delta$ F1\end{tabular} & \begin{tabular}[c]{@{}c@{}}cost (¢)\\ per pair\end{tabular} \\ \midrule
\multirow{4}{*}{gpt3.5-davinci-002} & domain-complex-forced-T                                                                     & 59.70 & 80.00  & 68.38 & 2.15                                               & 1.36                                                        \\
                                    & domain-simple-forced-T                                                                      & 72.34 & 68.00  & 70.10 & 3.87                                               & 1.29                                                        \\
                                    & general-complex-forced-T                                                                    & 43.10 & 100.00 & 60.24 & -5.99                                              & 1.40                                                        \\
                                    & general-simple-forced-T                                                                     & 65.50 & 80.00  & 70.18 & 3.95                                               & 1.29                                                        \\ \midrule
RoBERTa                             & \begin{tabular}[c]{@{}l@{}}fine-tuned on sampled\\ training set (2K pairs)\end{tabular}   & 85.99 & 80.00  & 82.72 & 16.49                                              & -                                                           \\
RoBERTa                             & \begin{tabular}[c]{@{}l@{}}fine-tuned on original\\ training set (20K pairs)\end{tabular} & 86.79 & 92.00  & 89.32 & 23.09                                              & -                                                           \\ \bottomrule
\end{tabular}%
}
\end{table}

\textbf{Baselines:} We compare the results of ChatGPT on our benchmark dataset to results of GPT-3 \textit{gpt3.5-davinci-002} which has been used by Narayan et al.~\cite{foundationalWrangleVLDB2022}, as well as to results for RoBERTa-base fine-tuned with different amounts of training data. A fine-tuned RoBERTa-base corresponds to the state-of-the-art entity matching system Ditto~\cite{liDeepEntityMatching2020} with all pre-processing and data augmentation options turned off. Results of the baseline methods are presented in Table \ref{tab:baselines}.
Using four of the prompt designs on the earlier gpt3.5-davinci-002 model shows that this model generally performs significantly worse compared to ChatGPT while having an about ten times higher cost per queried pair. The comparison with the fine-tuned RoBERTa-base baseline shows that ChatGPT in a zero-shot setting is able to reach a similar performance or even surpass RoBERTa fine-tuned with 2K training pairs. RoBERTa trained with 20K pairs is finally able to surpass most zero-shot prompts but its recall remains 6-8\% lower. The training data for both RoBERTa models contains product offers for the same products that are also part of the validation set, i.e. these products are considered in-distribution. It has been shown~\cite{peeters2023wdc} that such fine-tuned models experience a significant drop in performance when applied to pairs containing out-of-distribution products. The performance of ChatGPT in the zeroshot setup essentially corresponds to results on out-of-distribution data as no training happens, suggesting that ChatGPT is generally more robust concerning unseen products.


\section{In-Context Learning}
\label{Sec_InContext}

In the second set of experiments we analyse the impact of adding matching and non-matching product offer pairs as task demonstrations~\cite{liu-etal-2022-makes} to the prompts in order to help the model to understand and subsequently
perform the task correctly. We experiment with three different heuristics for selecting task demonstrations:

\begin{itemize}
    \item \textbf{Hand-picked}: Hand-picked demonstrations are a set of up to 10 matching and 10 non-matching product offer pairs which were hand-selected by a human domain expert from the pool of the training set. 
    \item \textbf{Random}: Demonstrations are drawn randomly from the labeled training set while making sure that they do not contain any of the products that are part of the product offer pair to be matched.
    \item \textbf{Related}: Related demonstrations are selected from the training set by calculating the Jaccard similarity between the pair to be matched and all positive and negative pairs in the training set. The resulting similarity lists are sorted and the most similar examples are selected.
\end{itemize}


In addition to the three selection heuristics, we also vary the amount of demonstration (shots) from 6 over 10 to 20 with an equal amount of positive and negative examples in order to evaluate the impact on performance and API cost. Due to their length, we do not provide examples of in-context prompts in this paper but refer the reader to the project github which contains all prompts. 

Table \ref{tab:in-context} shows the results of the in-context experiments. 
We compare the results to the zero-shot baseline of using domain-specific, complex language as well as forcing the model to answer with a simple "Yes" or "No" (see Section \ref{Sec_PromptDesign}).
For all three selection heuristics, providing 3 positive and 3 negative examples as task demonstrations leads to improvements over the zero-shot baselines of at least 2\% F1. Random demonstrations have the smallest impact with a maximum increase of 3.89\% (10 demonstrations) while the hand-picked demonstrations lead to an increase of up to 4.92\% (10 demonstrations) over the zeroshot baseline. Providing 20 related examples as demonstrations has the largest impact and improves the F1 score by nearly 8\% over the baseline. Across all in-context experiments, providing demonstrations consistently leads to an increase in precision while the recall decreases. This points to the model becoming more cautious when predicting positives. The more examples are provided the more pronounced this effect becomes. Providing task demonstrations is helpful in all cases as it provides the model with clear guidance on how the solutions to the task should look as well as patterns that correlate with the correct answer. The provision of related demonstrations increases this effect, as the model is steered towards patterns that are relevant for the decision at hand.

\begin{table}[]
\centering
\caption{Results of the in-context learning experiments and associated cost.}
\label{tab:in-context}
\resizebox{\textwidth}{!}{%
\begin{tabular}{@{}lcccccccc@{}}
\toprule
\begin{tabular}[c]{@{}l@{}}Selection heuristic\end{tabular} &
  \begin{tabular}[c]{@{}c@{}}Shots\end{tabular} &
  P &
  R &
  F1 &
  \begin{tabular}[c]{@{}c@{}}$\Delta$ F1\end{tabular} &
  \begin{tabular}[c]{@{}c@{}}Cost (¢)\\ per pair\end{tabular} &
  \begin{tabular}[c]{@{}c@{}}Cost\\ increase\end{tabular} &
  \begin{tabular}[c]{@{}c@{}}Cost increase\\ per $\Delta$ F1\end{tabular} \\ \midrule
ChatGPT-zeroshot                    & 0  & 71.01          & \textbf{98.00} & 82.35          & -                          & 0.14  & -                           & -      \\ \midrule
\multirow{3}{*}{ChatGPT-random}     & 6  & 78.33          & 94.00          & 85.45          & 3.10                       & 0.77  & 450\%                       & 145\%  \\
                                    & 10  & 79.66          & 94.00          & 86.24          & 3.89                       & 1.13  & 707\%                       & 182\%  \\
                                    & 20 & 78.95          & 90.00          & 84.11          & 1.76                       & 2.07  & 1379\%                      & 783\%  \\ \midrule
\multirow{3}{*}{ChatGPT-handpicked} & 6  & 76.19          & 96.00          & 84.86          & 2.51                       & 0.72  & 414\%                       & 165\%  \\
                                    & 10  & 80.00          & 96.00          & 87.27          & 4.92                       & 1.00  & 614\%                       & 125\%  \\
                                    & 20 & 79.66          & 94.00          & 86.24          & 3.89                       & 2.03  & 1350\%                      & 347\%  \\ \midrule
\multirow{3}{*}{ChatGPT-related}    & 6  & 80.36          & 90.00          & 84.91          & 2.56                       & 0.68  & 386\%                       & 151\%  \\
                                    & 10  & \textbf{89.58} & 86.00          & 87.76          & 5.41                       & 1.05  & 650\%                       & 120\%  \\
                                    & 20 & 88.46          & 92.00          & \textbf{90.20} & 7.85                       & 1.97  & 1307\%                      & 167\%  \\ \midrule
\multirow{2}{*}{GPT3.5-handpicked}  & 10  & 61.97          & 88.00          & 72.72          & \multicolumn{1}{r}{-9.63}  & 10.54 & \multicolumn{1}{r}{7429\%}  & 771\%  \\
                                    & 20 & 61.43          & 86.00          & 71.67          & \multicolumn{1}{r}{-10.68} & 19.71 & \multicolumn{1}{r}{13979\%} & 1309\% \\ \midrule
\multirow{2}{*}{GPT3.5-related}     & 10  & 67.69          & 88.00          & 76.52          & \multicolumn{1}{r}{-5.83}  & 10.04 & \multicolumn{1}{r}{7071\%}  & 1213\% \\
                                    & 20 & 61.43          & 86.00          & 71.67          & \multicolumn{1}{r}{-10.68} & 20.34 & \multicolumn{1}{r}{14429\%} & 1351\% \\ \bottomrule
\end{tabular}%
}
\end{table}

\textbf{Cost Analysis:} The performance gain resulting from task demonstrations comes with a sizable increase in API usage cost. While the zeroshot baseline prompt and model answer cost around 0.14¢ per matching decision, providing an additional 20 examples increases this cost by nearly 1400\% to 2¢ per decision. Breaking the increase in cost down to the increase per percentage point of F1 (see rightmost column in Table \ref{tab:in-context}) it becomes clear that providing related examples has the best price performance ratio, followed by hand-picked examples. This calculation does not factor in the cost of acquiring labeled pairs to be selected by the heuristics. As the handpicked demonstrations consists of only 20 labeled pairs, the labeling cost of this approach is significantly lower than the costs for the other two.



\section{Providing Matching Knowledge}
\label{Sec_MatchingKnwoledge}

\begin{figure}
\includegraphics[width=\textwidth]{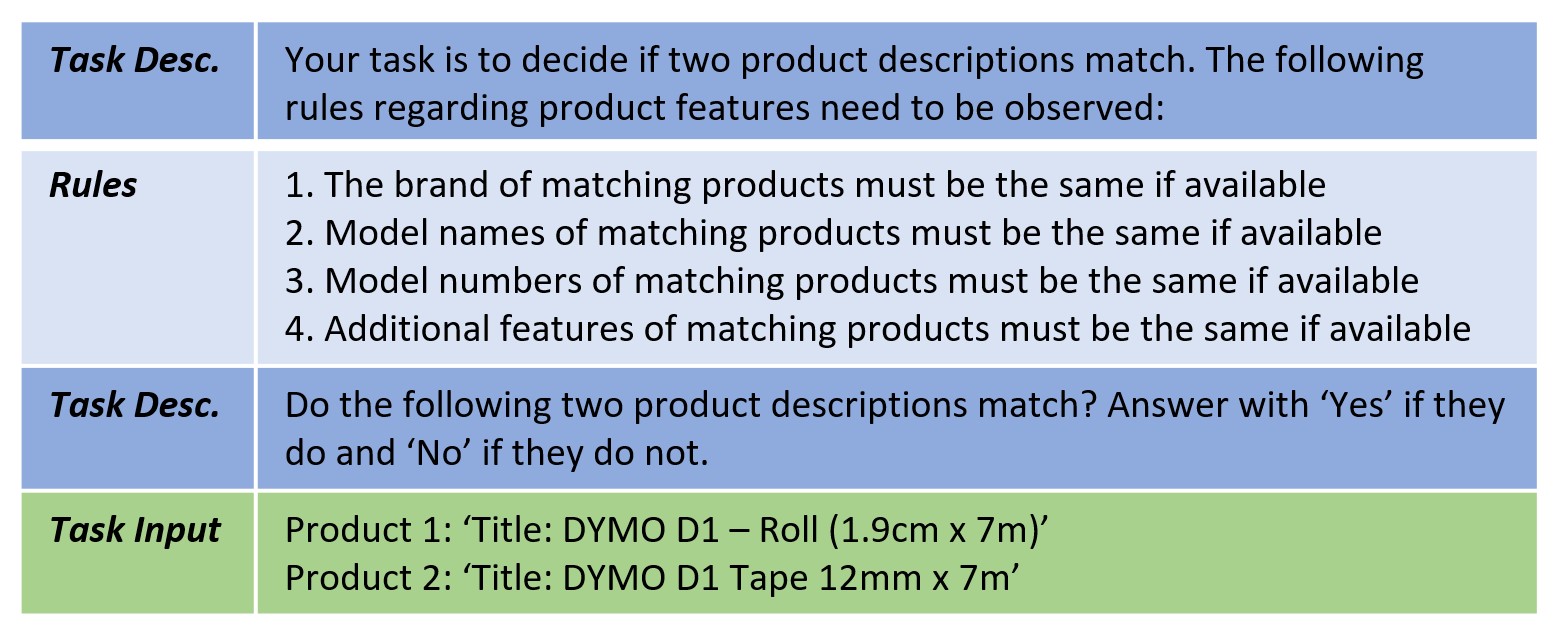}
\caption{Example of a prompt containing matching rules.} \label{fig:rules}
\end{figure}

The last set of experiments focuses on providing explicit matching knowledge in the form of natural language rules and asking the model to use these rules for its decisions. Asking ChatGPT to explain matching decisions revealed that ChatGPT is able to identify product features and corresponding feature values. Following this finding, we formulate explicit rules for a set of common product features as well as a general rule capturing any possible additional features. The natural language formulation of the rule set that we add to the prompts is shown in Figure \ref{fig:rules}. We experiment with using these rules in a zero-shot scenario as well as together with related demonstrations (see Section \ref{Sec_InContext}).

\begin{table}[]
\centering
\caption{Results of providing explicit matching knowledge to ChatGPT.}
\label{tab:rules}
\resizebox{\textwidth}{!}{%
\begin{tabular}{@{}lcccccccc@{}}
\toprule
\begin{tabular}[c]{@{}l@{}}Prompt\end{tabular} &
  \begin{tabular}[c]{@{}c@{}}Shots\end{tabular} &
  P &
  R &
  F1 &
  \begin{tabular}[c]{@{}c@{}}$\Delta$ F1\end{tabular} &
  \begin{tabular}[c]{@{}c@{}}Cost (¢)\\ per pair\end{tabular} &
  \begin{tabular}[c]{@{}c@{}}Cost\\ increase\end{tabular} &
  \begin{tabular}[c]{@{}c@{}}Cost increase\\ per $\Delta$ F1\end{tabular} \\ \midrule
ChatGPT-zeroshot                                                                      & 0  & 71.01          & \textbf{98.00} & 82.35          & -    & 0.14 & -      & -     \\ \midrule
\begin{tabular}[c]{@{}l@{}}ChatGPT-zeroshot\\ with rules\end{tabular}                 & 0  & 80.33          & \textbf{98.00} & 88.29          & 5.94 & 0.28 & 100\%  & 17\%  \\ \midrule
\multirow{3}{*}{ChatGPT-related}                                                      & 6  & 80.36          & 90.00          & 84.91          & 2.56 & 0.68 & 386\%  & 151\% \\
                                                                              & 10  & 89.58          & 86.00          & 87.76          & 5.41 & 1.05 & 650\%  & 120\% \\
                                                                              & 20 & 88.46          & 92.00          & \textbf{90.20} & 7.85 & 1.97 & 1307\% & 167\% \\ \midrule
\multirow{3}{*}{\begin{tabular}[c]{@{}l@{}}ChatGPT-related\\ with rules\end{tabular}} & 6  & 90.70          & 78.00          & 83.87          & 1.52 & 0.79 & 464\%  & 305\% \\
                                                                              & 10  & 90.91          & 80.00          & 85.11          & 2.76 & 1.17 & 736\%  & 267\% \\
                                                                              & 20 & \textbf{91.11} & 82.00          & 86.32          & 3.97 & 2.09 & 1393\% & 351\% \\ \bottomrule
\end{tabular}%
}
\end{table}

Table \ref{tab:rules} shows the results of adding matching rules to the prompts. Adding matching rules increases the zero-shot performance by 6\% F1 to 88.29\%. This performance is only 2\% lower than the performance that was reached by providing 20 related task demonstrations. Interestingly, providing the rules in a zero-shot setting does not negatively impact the recall, which remains at 98\%, but increases the precision of the model by nearly 10\%. Combining matching rules and 20 related demonstrations in a single prompt slightly further improves the precision but leads to a 10\% drop in recall and an overall lower F1. ChatGPT seems to be able to interpret matching rules and successfully applies them to improve matching results. Ultimately, task demonstrations and matching rules both serve the same purpose of guiding the model on how to match entities. Matching rules are more generic, while related demonstrations are rather product pair specific. This specificness might be the reason for the slightly higher performance. On the other hand, defining matching rules requires significantly less human effort compared to labeling a pool of examples for selecting related demonstrations. Adding explicit matching rules to prompts might thus be a promising approach for many real-world use cases.

\section{Conclusion}
\label{Sec_Conclusion}

We have demonstrated the impact of various prompt designs on the performance of ChatGPT on a challenging entity matching task. We have shown that the model can achieve competitive performance in a zero-shot setting compared to PLMs like RoBERTa which require to be fine-tuned using thousands of labeled examples. Due to the relative shortness of the prompts and the associated low API fees, using ChatGPT for entity matching can be considered a promising alternative to fine-tuned PLMs which require the costly collection and maintenance of large in-domain training sets. ChatGPT can further be considered more robust as it demonstrates competitive performance even in zero-shot settings while fine-tuned PLMs struggle with out-of-distribution entities which where not seen during training~\cite{peeters2023wdc}. The provision of task demonstrations further increases the performance, especially if the selected demonstrations are textually similar to the pair of entities to be matched. If closely related demonstrations are not available, providing randomly selected demonstrations also has a significant positive effect. The manual selection of just 20 demonstration pairs by a domain expert resulted in a strong positive effect and requires a significantly lower effort than having the domain expert label thousands of pairs for PLM-based matchers. Finally, providing the model with a set of explicit matching rules has a similar effect as providing textually related demonstrations. The provision of explicit, higher-level matching knowledge to LLMs via prompts is a promising direction for future research as it has the potential to significantly reduce the labeling effort required for achieving state-of-the-art entity matching results.

%
%
%
\bibliographystyle{splncs04}
\bibliography{main}
\end{document}